\documentclass[1p,sort&compress,number,12pt]{elsarticle}
\usepackage[utf8]{inputenc}
\usepackage[T1]{fontenc}
\usepackage{lmodern}
\usepackage[english]{babel}
\usepackage{enumitem}
\usepackage{amsmath}
\usepackage{amssymb}
\usepackage{graphicx}
\usepackage{wrapfig}
\usepackage{float}
\usepackage{xcolor}
\usepackage{algorithm}
\usepackage{algorithmic}
\usepackage{multirow}
\usepackage{bbm}

\usepackage{fancyhdr}
\begin{document}

\begin{frontmatter}
\title{Feature Selection for Regression Problems Based on the Morisita Estimator of Intrinsic Dimension}
\author{Jean GOLAY, Michael LEUENBERGER and Mikhail KANEVSKI}
\address{Institute of Earth Surface Dynamics, Faculty of Geosciences and Environment, University of Lausanne, 1015 Lausanne, Switzerland. Email: jean.golay@unil.ch.}

\begin{abstract}
Data acquisition, storage and management have been improved, while the key factors of many phenomena are not well known. Consequently, irrelevant and redundant features artificially increase the size of datasets, which complicates learning tasks, such as regression. To address this problem, feature selection methods have been proposed. This paper introduces a new supervised filter based on the Morisita estimator of intrinsic dimension. It can identify relevant features and distinguish between redundant and irrelevant information. Besides, it offers a clear graphical representation of the results, and it can be easily implemented in different programming languages. Comprehensive numerical experiments are conducted using simulated datasets characterized by different levels of complexity, sample size and noise. The suggested algorithm is also successfully tested on a selection of real world applications and compared with RReliefF using extreme learning machine. In addition, a new measure of feature relevance is presented and discussed.
\end{abstract}

\begin{keyword}
Feature selection \sep Intrinsic dimension \sep Morisita index \sep Measure of relevance \sep Data mining 
\end{keyword}
\end{frontmatter}

\section{Introduction}
In data mining, it is often not known a priori what features (or input variables \footnote{In this paper, the term ``feature'' is used as a synonym for ``input variable''.}) are truly necessary to capture the main characteristics of a studied phenomenon. This lack of knowledge implies that many of the considered features are irrelevant or redundant. They artificially increase the dimension $E$ of the Euclidean space $\mathbb{R}^E$ in which the data points are embedded ($E$ equals the number of input and output variables under consideration). This is a serious matter, since fast improvements in data acquisition, storage and management cause the number of redundant and irrelevant features to increase. As a consequence, the interpretation of the results becomes more complicated and, unless the sample size $N$ grows exponentially with $E$, the curse of dimensionality \cite{Bell61} may reduce the overall accuracy yielded by any learning algorithm. Besides, large $N$ and $E$ are also difficult to deal with because of computer performance limitations.

In regression and classification, these issues are often addressed by implementing supervised feature selection methods \cite{Guy03,Guy06,Ghe10,Zen15}. Such methods can be broadly subdivided into filter (e.g. RReliefF \cite{Robnik03}, mRMR \cite{Pen05} and CFS \cite{Hall00}), wrapper \cite{Koh97,Leu14} and embedded methods (e.g. the Lasso \cite{Tib96} and random forest \cite{Bre01}). Filters rank features, or subsets of features, according to a relevance measure independently of any predictive model, while wrappers use an evaluation criterion involving a learning machine. Both approaches can be used with search strategies, since an exhaustive exploration of the $2^{\# Feat.}-1$ models (all the possible combinations of features) is often computationally intractable. Greedy strategies \cite{Cot01,Col03}, such as Sequential Forward Selection (SFS) \cite{Whit71}, can be distinguished from stochastic ones (e.g. simulated annealing \cite{Mei06,Kirk83} and ant colony optimization \cite{Ta15,Do92}). Regarding the embedded methods, the feature selection is a by-product of a training procedure. It can be achieved by the addition of constraints in the cost function of a predictive model (e.g. the Lasso \cite{Tib96}), or it can be more specific to a given algorithm (e.g. random forest \cite{Bre01} and adaptive general regression neural networks \cite{Rob13}).

The present paper\footnote{The main idea of this paper was partly presented at the 23rd symposium on artificial neural networks, computational intelligence and machine learning (ESANN2015) \cite{Go15Esann}.} deals with a new SFS filter algorithm. It relies on the idea that, although data points are embedded in $E$-dimensional spaces, they often reside on lower $M$-dimensional manifolds \cite{Cama03,LeeVer07,Cama16}. The value $M$ ($\leq E$) is called Intrinsic Dimension (ID), and it can be estimated using the Morisita estimator of ID \cite{Go15} which is closely related to the fractal theory. The proposed filter algorithm is supervised, designed for regression problems and based on this new ID estimator. It also keeps the simplicity of the Fractal Dimension Reduction (FDR) algorithm introduced in \cite{Trai00}. Finally, the results show the ability of the new filter to capture non-linear relationships and to effectively identify both redundant and irrelevant information.

The paper is organized as follows. Section \ref{overview} reviews previous work on ID-based feature selection approaches. The Morisita estimator of ID is shortly presented in Section \ref{Mindex} (for the completeness of the paper). Section \ref{MBFR} introduces the Morisita-based filter, and Section \ref{Exp_art} is devoted to numerical experiments conducted on simulated data of varying complexity. In Section \ref{Exp_real}, real world applications from publicly accessible repositories are presented, and a comparison with a benchmark algorithm, RReliefF \cite{Robnik03}, is carried out using Extreme Learning Machine (ELM) \cite{Hua06}. Finally, conclusions are drawn in the last section with a special emphasis on future challenges and applications. 

\section{Related Work}\label{overview}
The concept of ID can be extended to the more general case where the data ID may be a non-integer dimension $D$ \cite{Trai00,Ek11,LeeVer07}. The value $D$ is estimated by using fractal-based methods which have been presented in \cite{Cama16,LeeVer07,Go14} and successfully implemented in various fields, such as physics \cite{Grass833}, cosmology \cite{Bor93}, meteorology \cite{lov87} and pattern recognition \cite{Hua94,Xu15}. These methods rely on well-known fractal dimensions (e.g. the box-counting dimension \cite{Mand83,Ott93}, the correlation dimension \cite{Grass833} and R\'{e}nyi's dimensions of $q$th order \cite{Hent83}), and they can be used in feature selection \cite{Trai00,Mo12} and dimensionality reduction \cite{LeeVer07} to detect dependencies between variables (or features).

Traina et al. \cite{Trai00,Trai10} have opened up new prospects for the effective use of ID estimation in data mining by introducing the Fractal Dimension Reduction (FDR) algorithm. FDR executes an unsupervised procedure of feature selection aiming to remove from a dataset all the redundant variables. The fundamental idea is that fully redundant variables do not contribute to the value of the data ID.

This idea can be illustrated by sampling two uniformly distributed variables $V_1$ and $V_2$. If they are independent, which means that they are not redundant, one has that:\begin{equation}\label{Eq_concept}
ID(V_1,V_2)\approx ID(V_1)+ID(V_2)\approx 1+1=2
\end{equation}where $ID(\cdot)$ denotes the ID of a dataset. It indicates that both $V_1$ and $V_2$ contribute to increasing the value of $ID(V_1,V_2)$ by about $1$, which is, by construction, equal to the ID of each variable (i.e. $ID(V_1)$ and $ID(V_2)$). Conversely, the removal of either $V_1$ or $V_2$ would lead to a reduction in the data ID from about 2 (i.e. the dimension of the data space) to $1$ (i.e. the ID of a single variable) and information would be irreparably lost. In contrast, if $V_1$ and $V_2$ are fully redundant with each other (e.g. $V_2 = V_1$), one has that: \begin{equation}
ID(V_1,V_2)\approx ID(V_1)\approx ID(V_2)\approx 1
\end{equation}where the ID of the full dataset is approximately equal to the topological dimension of a smooth line. This means that the contribution of only one variable is enough to reach the value of $ID(V_1,V_2)$ and the remaining one can be disregarded without losing any information.

Based on these considerations, the FDR algorithm removes the redundant variables from a dataset by implementing a Sequential Backward Elimination (SBE) strategy \cite{Cot01}. Besides, it uses R\'{e}nyi's dimension of order $q=2$, $D_2$, for the ID estimation. Following the same principles, De Sousa et al. \cite{Sousa07} examined additional developments to FDR and presented a new algorithm for identifying subgroups of correlated variables.

FDR is designed to carry out unsupervised tasks, and it is not able to distinguish between variables that are relevant to a learning process and those that are irrelevant. The reason is that such variables can all contribute to the data ID. For instance, in Equation \ref{Eq_concept}, $V_1$ could be regarded as irrelevant to the learning of $V_2$, but it would be selected by FDR because it makes the data ID increase by about $1$. Consequently, different studies were carried out to adapt FDR to supervised learning. Lee et al. \cite{Lee06} suggested decoupling the relevance and redundancy analysis. Following the same idea, Pham et al. \cite{Pham09} used mutual information to identify irrelevant features and combined the results with those of FDR. Finally, Mo and Huang \cite{Mo12} developed an advanced algorithm to detect both redundant and irrelevant information in a single step. Their algorithm follows a SBE search strategy and relies on the correlation dimension, $df_{cor}$, for the estimation of the data ID.

The filter algorithm suggested in the present paper is designed in such a way that it combines the advantages of both FDR and Mo's algorithm: it can deal with non-linear dependencies, it does not rely on any user-defined threshold, it can discriminate between redundant and irrelevant information, and the results can be easily summarized in informative plots. Moreover, it can cope with high-dimensional datasets thanks to its SFS search strategy, and it uses the Morisita estimator of ID which was shown to yield comparable or better results than $D_2$ and $df_{cor}$ \cite{Go15}.

\section{The Morisita Estimator of Intrinsic Dimension}\label{Mindex}
The Morisita estimator of ID, $M_m$, has been recently introduced \cite{Go15}. It is a fractal-based ID estimator derived from the multipoint Morisita index $I_{m,\delta}$ \cite{Go14,Hul90} (named after Masaaki Morisita who proposed the first version of the index to study the spatial clustering of ecological data \cite{Mori59}). $I_{m,\delta}$ is computed by superimposing an $E$-dimensional grid of $Q$ quadrats of diagonal size $\delta$ onto the data points. It measures how many times more likely it is that $m$ ($m\geq 2$) randomly selected points will be from the same quadrat than it would be if all the $N$ points of the studied dataset were distributed at random (i.e. according to a random distribution generated from a Poisson process). The formula is the following:\begin{equation}
I_{m,\delta}=Q^{m-1}\frac{\sum_{i=1}^Q n_i(n_i-1)(n_i-2) \dotsm (n_i-m+1)}{N(N-1)(N-2) \dotsm (N-m+1)}
\end{equation}where $n_i$ is the number of points in the $i^{th}$ quadrat. For a fixed value of $m$, $I_{m,\delta}$ is calculated for a chosen scale range. If a dataset approximates a fractal behavior (i.e. is self-similar) within this range, the relationship of the plot relating $\log{(I_{m,\delta})}$ to $\log{(1/\delta)}$ is linear, and the slope of the regression line is defined as the Morisita slope $S_m$. Finally, $M_m$ is expressed as:\begin{equation}\label{Eq_mindid}M_m = E - \left( \frac{S_{m}}{m-1}\right).\end{equation}In practice, each variable is rescaled to the $[0,1]$ interval (so is the grid), and $\delta$ can be replaced with the quadrat edge length $\ell$, with $\ell^{-1}$ being simply the number of quadrats along each axis of the data space. Then a set of $R$ values of $\ell$ (or $\ell^{-1}$) is chosen so that it captures the linear part of the log-log plot. In the rest of this paper, only $M_{m=2}$ will be used, and it will be computed with an algorithm called Morisita INDex for Intrinsic Dimension estimation (MINDID) \cite{Go15} whose complexity is $\mathcal{O}(N*E*R)$.

\section{The Morisita-based Filter for Regression Problems}\label{MBFR}
The Morisita-Based Filter for Regression problems (MBFR) relies on three observations following from the work by Traina et al. \cite{Trai00}, De Sousa et al. \cite{Sousa07} and Mo and Huang \cite{Mo12}:\begin{enumerate}
\item Given an output variable $Y$ generated from $k$ relevant and non-redundant input variables $X_1,\ldots ,X_k$, one has that:
\begin{equation} 
ID(X_1,\ldots ,X_k,Y)-ID(X_1,\ldots ,X_k)\approx 0 
\end{equation}where $ID(\cdot)$ denotes the (possibly non-integer) ID of a dataset.
\item Given $i$ irrelevant input variables $I_1,\ldots ,I_i$ completely independent of Y, one has that:
\begin{equation} 
ID(I_1,\ldots ,I_i,Y)-ID(I_1,\ldots ,I_i)\approx ID(Y) 
\end{equation}
\item Given a randomly selected subset of $\left\lbrace X_1,\ldots ,X_k\right\rbrace$ of size $r$ with $1\leq r<k$ and $k\geq 2$, $j$ redundant input variables $J_1,\ldots ,J_j$ related to some or all $X_1,\ldots ,X_r$ and all the $i$ irrelevant input variables $I_1,\ldots ,I_i$, one has that:
\begin{equation}
\begin{aligned} 
&ID(X_1,\ldots ,X_r,J_1,\ldots ,J_j,I_1,\ldots ,I_i,Y)\\
-&ID(X_1,\ldots ,X_r,J_1,\ldots ,J_j,I_1,\ldots ,I_i)\approx H 
\end{aligned}\end{equation}where $H\in \left]0,ID(Y)\right[$ and $H$ decreases to $0$ as $r$ increases to $k$.    
\end{enumerate}

\begin{algorithm}[t]
\caption{MBFR}\label{mbfr_algo}
\textbf{INPUT:} 

A dataset $A$ with $E-1$ features $F_{1,\ldots ,E-1}$ and one output variable $Y$.
 
A vector $L$ of values $\ell^{-1}$.
 
An integer $C$ ($\leq E-1$) indicating the number of steps of the SFS to be performed.
 
Two empty vectors of length $C$: $SelF$ and $DissF$ for storing, respectively, the names of the selected features and the dissimilarity values.
 
An empty matrix $Z$ for storing the selected features.

\textbf{OUTPUT:} $SelF$ and $DissF$. 
\begin{algorithmic}[1]
\STATE Rescale each feature and $Y$ to $[0,1]$.
\FOR{$i = 1 \ \TO \ C$}
\FOR{$j = 1 \ \TO \ (E-i)$}
\STATE $\widehat{Diss}(Z,F_j,Y)=M_2(Z,F_j,Y)-M_2(Z,F_j)$ (MINDID used with $L$)
\ENDFOR
\STATE Store in $SelF[i]$ the name of the $F_j$ yielding the lowest value of $\widehat{Diss}$. 
\STATE Store this value of $\widehat{Diss}$ in $DissF[i]$. 
\STATE Remove the corresponding $F_j$ from $A$ and add it into $Z$.
\ENDFOR\\
\end{algorithmic}
\end{algorithm} 

The difference \begin{equation}\label{diss_equ}Diss(F,Y):=ID(F,Y)-ID(F)\end{equation}can thus be suggested as a way of measuring the dissimilarity (i.e the independence) between $Y$ and a set $F$ of features (e.g. $F=\lbrace X_1,X_2,J_1,I_1\rbrace$), among which only the relevant ones (i.e. the non-redundant features on which $Y$ depends) contribute to reducing the dissimilarity. Based on that idea, MBFR (see Algorithm \ref{mbfr_algo}) aims at retrieving the relevant features available in a dataset by sorting each subset of variables according to its dissimilarity with $Y$. MBFR implements a SFS search strategy and relies on the Morisita estimator of ID and the MINDID algorithm \cite{Go15} to estimate $Diss$:\begin{equation}\label{diss_equ_est}\widehat{Diss}(F,Y):=M_2(F,Y)-M_2(F).\end{equation}In terms of time complexity, the algorithm is linear on $N$ and $R$, but its bottleneck is the SFS search strategy which is quadratic on $E$. In spite of this limitation, the execution time of MBFR remains competitive as shown in Section \ref{Exp_real}. It can also be significantly reduced by setting $C$ (i.e. the number of steps of the SFS procedure to be performed) to a small value. For instance, if $Diss$ is likely to reach its minimum value after only a few SFS steps because of many redundant and irrelevant features, $C$ can be set to a value substantially lower than $E-1$.

For ease of comparison, the coefficient of dimensional relevance, $DR(F,Y)$, can be introduced. It is defined as:\begin{equation}\label{Eq_DR} 
DR(F,Y):=1-\frac{Diss(F,Y)}{ID(Y)}=1-\frac{ID(F,Y)-ID(F)}{ID(Y)} 
\end{equation}and it can be computed using the Morisita estimator of ID $M_2$. In the same way as $Diss(F,Y)$, $DR(F,Y)$ is able to capture both linear and non-linear relationships between an input and an output space. Besides, it lies between $0$ and $1$. If the target (or output) variable $Y$ can be completely explained by the considered features $F$, $DR(F,Y) = 1$. On the contrary, if all the available features are irrelevant, $DR(F,Y) = 0$, and in-between, the closer it is to $1$, the greater the predictive power of $F$.

\begin{figure}
\centering
\includegraphics[width=\linewidth]{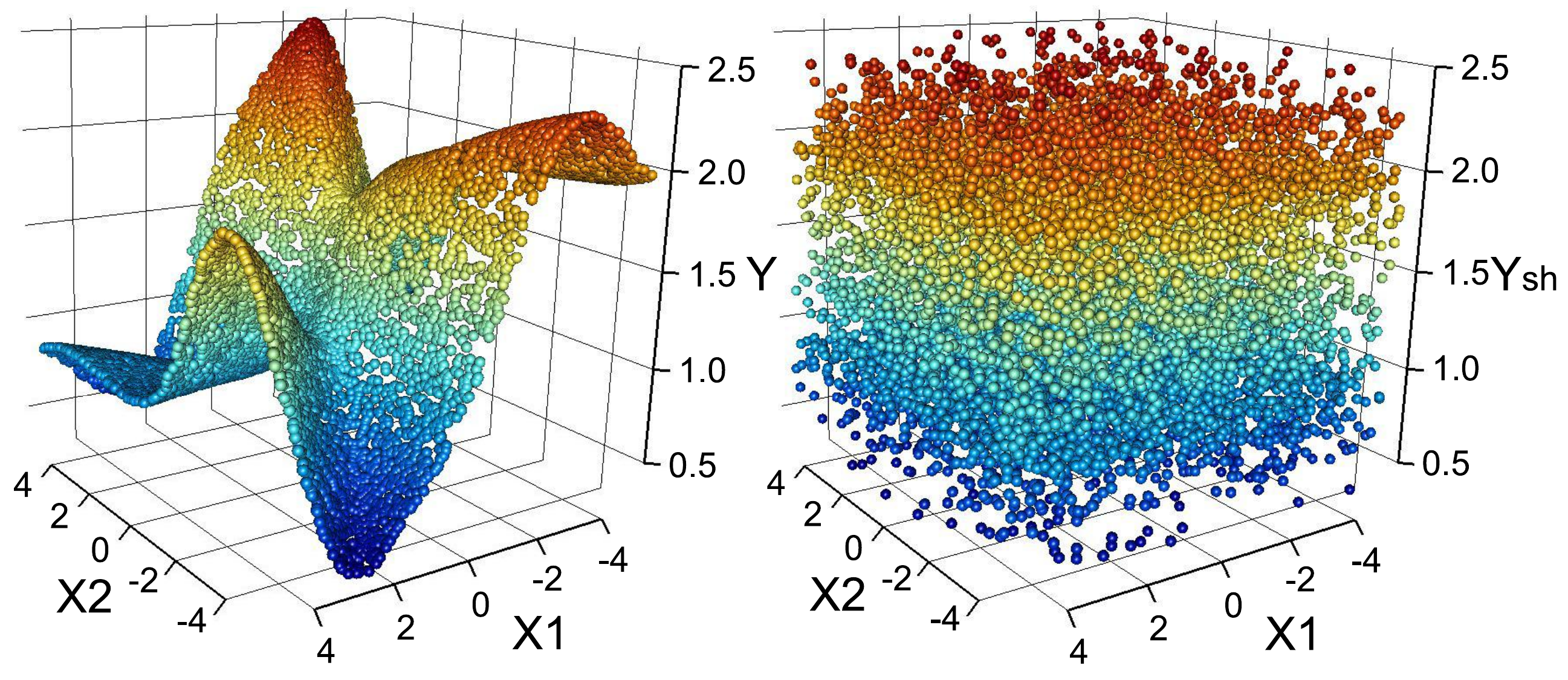}
\caption{(left) The functional relationship between the output variable $Y$ and the relevant features $X_1$ and $X_2$ of the butterfly dataset; (right) Shuffling of the output variable $Y$.}\label{Fig_butterVisual}
\end{figure}

\begin{table}[t]
\centering 
\footnotesize
\begin{tabular}{cccc}
\hline 
$j$& $\omega_{1,j}$ & $\omega_{2,j}$ & $\beta_j$ \\
\hline 
\hline
$1$&  $0.6655$ & $0.8939$ & $1.3446$\\   
$2$&  $1.2611$ & $-0.3512$ & $-0.0115$\\
$3$&  $0.3961$ & $-1.7827$ & $1.2770$\\
$4$&  $-1.7065$ & $-0.5297$ & $0.5962$\\ 
$5$&  $0.8807$ & $1.9574$ & $-0.8530$\\ 
$6$&  $1.8260$ & $0.7962$ & $-0.7290$\\ 
$7$&  $1.3400$ & $1.5001$ & $1.2339$ \\ 
$8$&  $1.2919$ & $-0.4462$ & $0.1186$\\ 
$9$&  $-1.3902$ & $1.6856$ & $0.5277$ \\ 
$10$& $0.0743$ & $1.5625$ & $-0.6952$ \\  
\hline 
\end{tabular}
\caption{Weights used in the construction of the butterfly dataset.}
\label{T6}
\end{table}

\begin{figure}[t!]
\centering
\includegraphics[width=\linewidth]{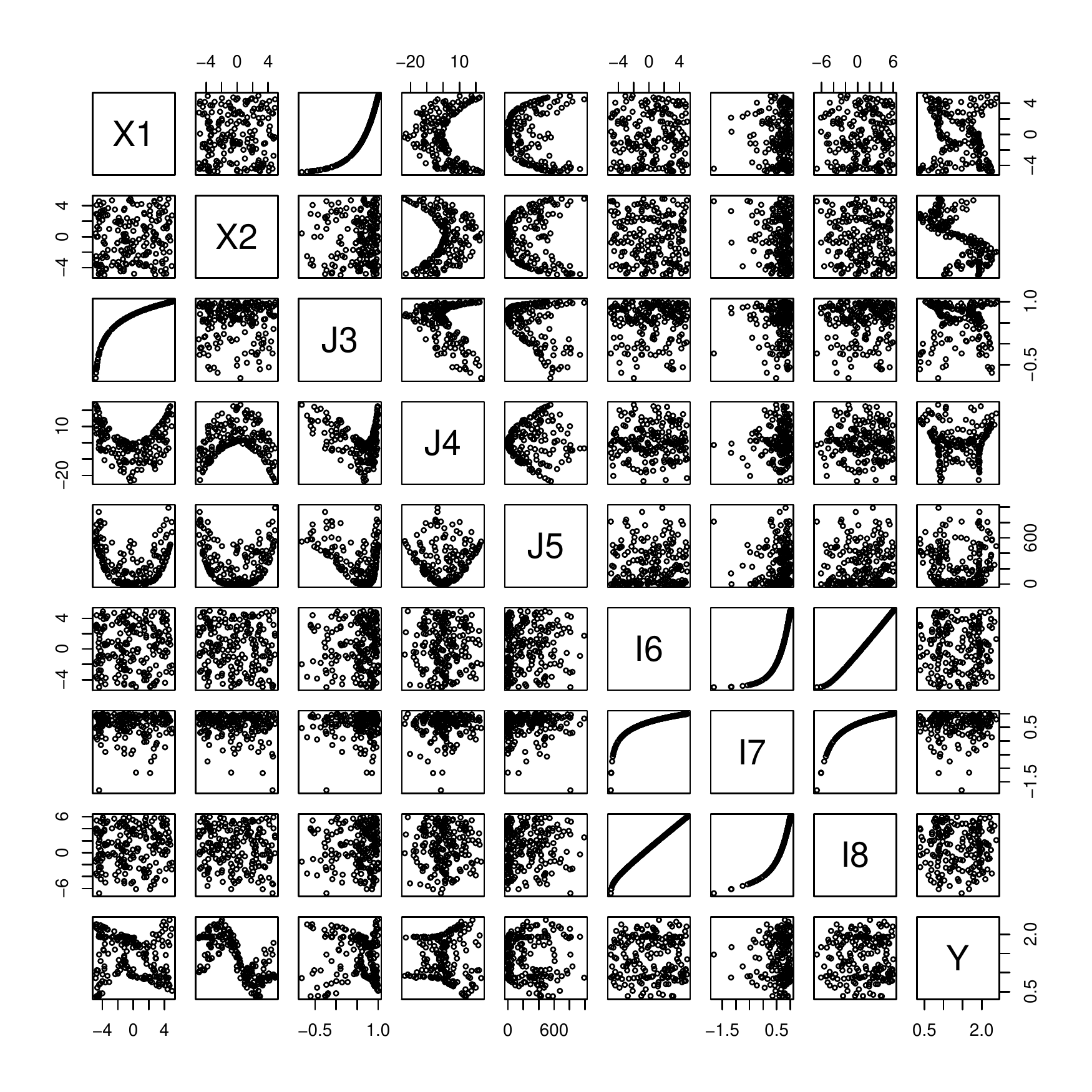}
\caption{Scatterplot matrix of the butterfly dataset for one simulation.}\label{Fig_scplot}
\end{figure}

\section{Experimental Study Using Simulated Data}\label{Exp_art}
In this section, the MBFR algorithm is assessed by means of two simulated datasets (see Subsection \ref{sim_data}), and its overall performance is carefully examined through a set of questions around which the subsections are organized: \begin{itemize}[noitemsep]
\item Question 1: How does sample size affect MBFR (see Subsection \ref{samplesize})?
\item Question 2: How does the complexity of data manifolds affect MBFR (see Subsection \ref{manifold}) ?
\item Question 3: How can MBFR help distinguish between redundant and irrelevant information (see Subsection \ref{info})?
\item Question 4: How does MBFR respond to a (partial) lack of relevant information (see Subsection \ref{no_info})?
\item Question 5: How does MBFR respond to the presence of noise in data (see Subsection \ref{noise})?
\end{itemize}
Notice also that the R environment \cite{Rcran} was used to implement the MBFR algorithm and to carry out the experiments.

\subsection{Simulated Datasets}\label{sim_data}
Two simulated datasets were used: the butterfly and Friedman datasets.
\begin{enumerate}
\item The butterfly dataset\footnote{It can be downloaded from: https://sites.google.com/site/jeangolayresearch/.} (see Figure \ref{Fig_butterVisual}): An output variable $Y$ is generated from two uniformly distributed input variables $X_1,X_2 \in \left] -5,5\right[$ by using an Artificial Neural Network (ANN) consisting of one hidden layer of $10$ neurons. It can be expressed as:\begin{equation}\label{equ_butter}
Y = \left[ \sum_{j=1}^{10}\beta_j sig(X_1 \omega_{1,j} + X_2 \omega_{2,j})\right]  + \varepsilon 
\end{equation}where $\omega_{1,j}$ and $\omega_{2,j}$ are the weights connecting the input variables to the $j^{th}$ neuron, $sig\left(x \right) : \mathbb{R}\rightarrow \mathbb{R}$ is a sigmoid transfer function, $\beta_j$ is the weight between the $j^{th}$ neuron and the output layer, and $\varepsilon$ is a Gaussian noise with zero mean and varying standard deviation (by default, it is set to $0.00$). The exact weights used in the construction of the dataset are given in Table \ref{T6}. Moreover, the addition of three redundant ($J$) and three irrelevant ($I$) variables is also made to complete the input space: $J_3=log_{10}{(X_1+5)}$, $J_4 = X_1^2-X_2^2$, $J_5 = X_1^4-X_2^4$, a uniformly distributed variable $I_6\in \left]-5,5\right[$, $I_7=log_{10}{(I_6+5)}$ and $I_{8}=I_6+I_7$. Finally, the butterfly dataset is generated by random sampling of $X_1$, $X_2$ and $I_6$. In this paper, different sample sizes were considered: $N=1000,2000,10000,20000$. Figure \ref{Fig_scplot} shows the scatterplot matrix of the full dataset for one simulation. The matrix highlights that the features were constructed so that the butterfly data not only contain linear relationships, but also a wide range of non-linear redundancies.
\item The Friedman dataset: this dataset uses a function suggested in \cite{Fri91} to test the ability of Multivariate Adaptive Regression Splines (MARS) models to uncover structures in data. The output $Y$ is given by:\begin{equation}\begin{aligned}Y = &10 \sin(\pi X_1 X_2)  +  20(X_3-0.5)^2  +\\ 
                                   &10 X_4  +  5 X_5  +  \varepsilon  \end{aligned}\end{equation}where $X_1$, $X_2$, $X_3$, $X_4$ and $X_5$ are i.i.d. variables following a uniform distribution $\textit{Unif}(0,1)$, and $\varepsilon$ is a Gaussian random noise with zero mean and unit variance. The input space is then completed by the addition of five irrelevant variables ($I$) following the same uniform distribution: $I_6$, $I_7$, $I_8$, $I_9$ and $I_{10} \sim \textit{Unif}(0,1)$. Finally, the Friedman dataset is produced by randomly sampling $N$ points from the inputs. In this paper, the sample size was set to $N=40000$ in accordance with the version of the dataset available on the Regression website \cite{Tor}. 
\end{enumerate}

The butterfly and Friedman datasets are characterized by non-linear structures, and their input spaces contain extra variables (i.e. redundant and irrelevant variables) that can be removed without affecting the learning of the target $Y$. In the following subsections, MBFR will be subjected to a battery of tests to highlight its ability to select the relevant variables ($X$) and to remove the irrelevant ($I$) and redundant ($J$) ones. Additional experiments will consider shuffled data to examine the response of the algorithm to a complete absence of structure. In parallel, the variability of the results will be examined by means of Monte Carlo simulations: for each experiment, many simulations of the datasets will be generated by repeated random sampling of the input variables.

Notice also that the way the two datasets are constructed leads to the distinction between the data manifolds and the manifolds of the simulated phenomena. The former are built using all the variables (including the output variable), while the latter (referred to as the Friedman and butterfly manifolds) do not involve the irrelevant features.
 
Finally, from the perspective of MBFR, a dataset is fully characterized by the integer values of $\ell^{-1}$. For the butterfly and Friedman datasets, these values were respectively set to $\lbrace5,6,\ldots,20\rbrace$ and $\lbrace1,2,\ldots,6\rbrace$. The two sets were chosen, so that, within their bounds, the relationship between $\log{(I_{m=2,\ell^{-1}})}$ and $\log{(\ell^{-1})}$ was linear. Notice that the upper bound of the second set is lower than that of the first one. This partially follows from the fact that the Friedman dataset has the greatest ID causing the data points to be sparsely distributed inside the data space. As a consequence, beyond $\ell^{-1}=5$, the probability of drawing two points from the same cell is rather low, while it is possible to use values of $\ell^{-1}$ up to $20$ in the case of the butterfly dataset.

\begin{table}
\centering 
\footnotesize
\begin{tabular}{cccc}
\hline 
$N$    &  First Two Features (Occurrences) &  $mean(DR)$  & $sd(DR)$\\ 
\hline 
\hline
1000     & $X_1$,$X_2$ ($99$); $X_2$,$X_1$ ($1$)  & $0.97$ &  $0.02$\\   
2000     &           $X_1$,$X_2$ ($100$)          & $0.97$ &  $0.02$\\ 
10000    &           $X_1$,$X_2$ ($100$)          & $0.97$ &  $0.01$\\ 
\hline 
\end{tabular}
\caption{The first two features selected by MBFR when applied successively to $100$ simulations of the butterfly dataset for different sample sizes. The mean values and the standard deviations of $DR$ are also provided.}
\label{T7}
\end{table}

\begin{figure}[t]
\centering
\includegraphics[width=\linewidth]{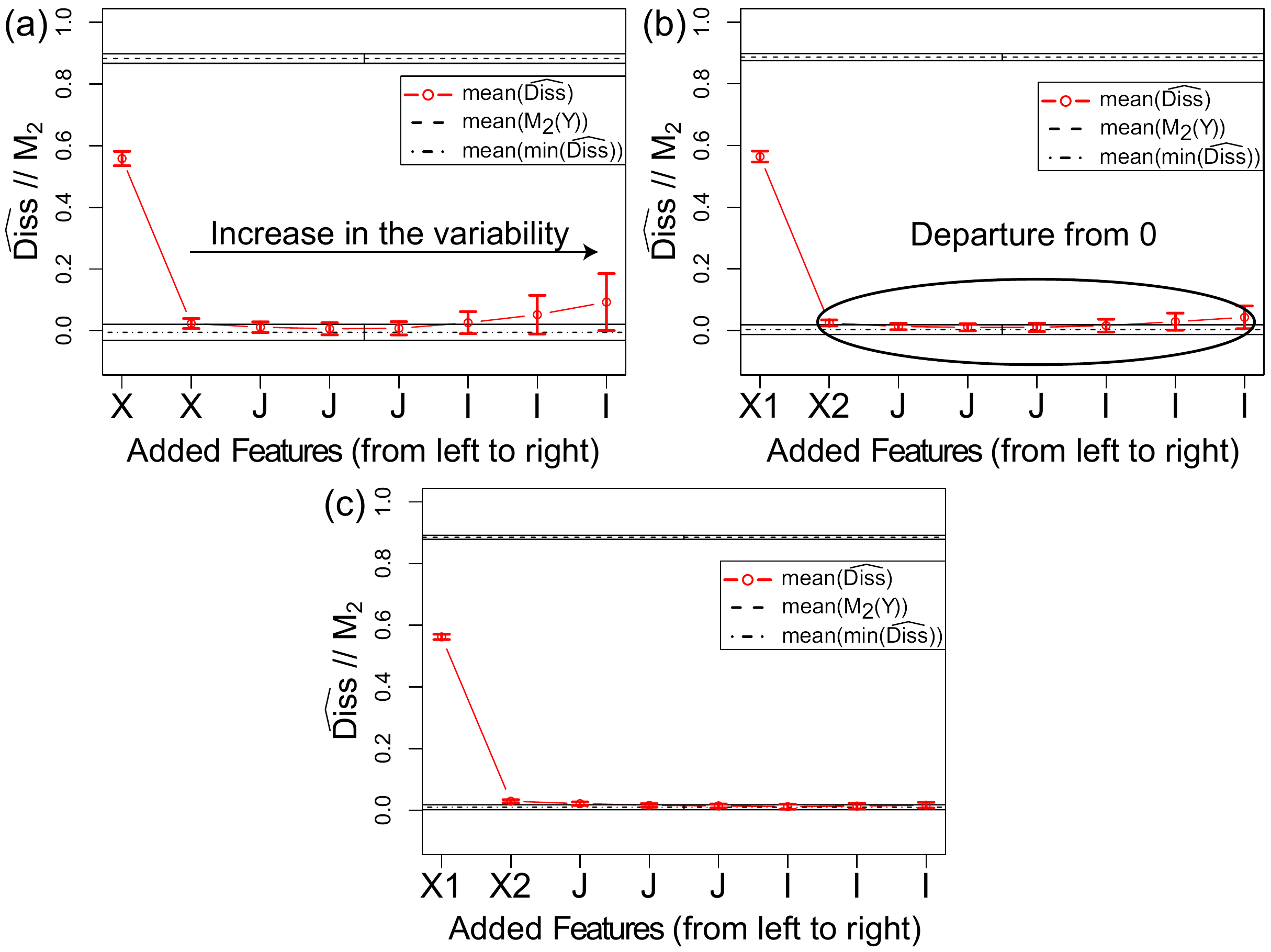}
\caption{Results of the application of MBFR to 100 simulations of the butterfly data with (a) $N=1000$, (b) $N=2000$ and (c) $N=10000$. Notice that the names of the features were shortened to $X$, $J$ and $I$ if the way they were ordered by the SFS search strategy changed between the simulations.}\label{Fig_nbr_points}
\end{figure}

\subsection{Sample Size}\label{samplesize}
MBFR was applied to the butterfly dataset. Three sample sizes were successively considered ($N=1000$, $2000$, $10000$), and for each of them, $100$ simulations of the data were produced.

Table \ref{T7} gives the first two input variables selected by the algorithm and indicates the number of times they were selected first over the simulations. Regardless of the sample size, MBFR always identifies $X_1$ and $X_2$ as the most relevant features, although their order can be reversed for $N=1000$. Besides, the predictive power of these two features was assessed using the coefficient of dimensional relevance $DR$. Table \ref{T7} provides the mean values of $DR$ over the simulations as well as the corresponding standard deviations (sd). The means are close to $1$, which implies that $X_1$ and $X_2$ convey most of the information contained in the dataset, and this is in agreement with the data construction: $X_1$ and $X_2$ are sufficient to explain $Y$, whereas the other features are not necessary or even useless. Moreover, the SFS search strategy enables MBFR to select the most relevant features by exploring a rather low-dimensional space. Consequently, the variability of $DR$ remains roughly constant over the sample sizes, and the standard deviations do not exceed $0.02$. 

To explore further the potential of the MBFR algorithm, a new series of $100$ simulations were generated. For ease of comparison with the next subsections, a constraint was imposed that for each simulation the redundant variables ($J$) had to be selected by MBFR before the irrelevant ones ($I$). The results are plotted in Figure \ref{Fig_nbr_points}.

The red dots indicate the mean dissimilarity values that are computed, over the simulations, by adding to the input space the features appearing on the horizontal axis. In addition, the red bars are the corresponding standard deviations. The features are progressively selected from left to right according to the SFS search strategy of MBFR, and the names of the redundant and irrelevant features were shortened to the letters $J$ and $I$ because they happened to switch position between the simulations. For the same reasons, $X_1$ and $X_2$ were replaced with the letter $X$ for $N=1000$. Furthermore, in each plot, two additional values are provided: the mean ID estimate of the target variable $Y$ (i.e. $mean\left(M_2\left(Y \right)\right)$) and the mean of the minimum dissimilarity (i.e. $mean\left(min\left(\widehat{Diss} \right)\right)$). The standard deviations of the two values are indicated using the black stripes.

For each sample size, $X_1$ and $X_2$ are easily identified as the two relevant features, since they contribute to reducing the dissimilarity from $M_2(Y)$ to about $0$ and a clear cut-off point is visible. However, as the number of points is reduced, the variability of the dissimilarity estimates increases. It does not question the potential of the algorithm for feature selection, but it emphasizes two aspects of its implementation: (1) the progressive increase in the variability as more features are added, and (2) the departure from $0$ of the mean dissimilarity estimates after the addition of the second relevant feature (see Figure \ref{Fig_nbr_points}). These two aspects will be addressed in the next subsection.

\subsection{Complexity of data manifolds}\label{manifold}
Lower sample sizes highlight that the variability of the dissimilarity values progressively increase as more features are picked out (see Figure \ref{Fig_nbr_points}). This response of MBFR is partly due to the presence of the relevant and irrelevant features which amplifies the data ID during the SFS procedure. But it is also related to the non-linear constructions of these features that affect the ID estimates by altering the point clustering on the data manifold. 

\begin{figure}[ht]
\centering
\includegraphics[width=\linewidth]{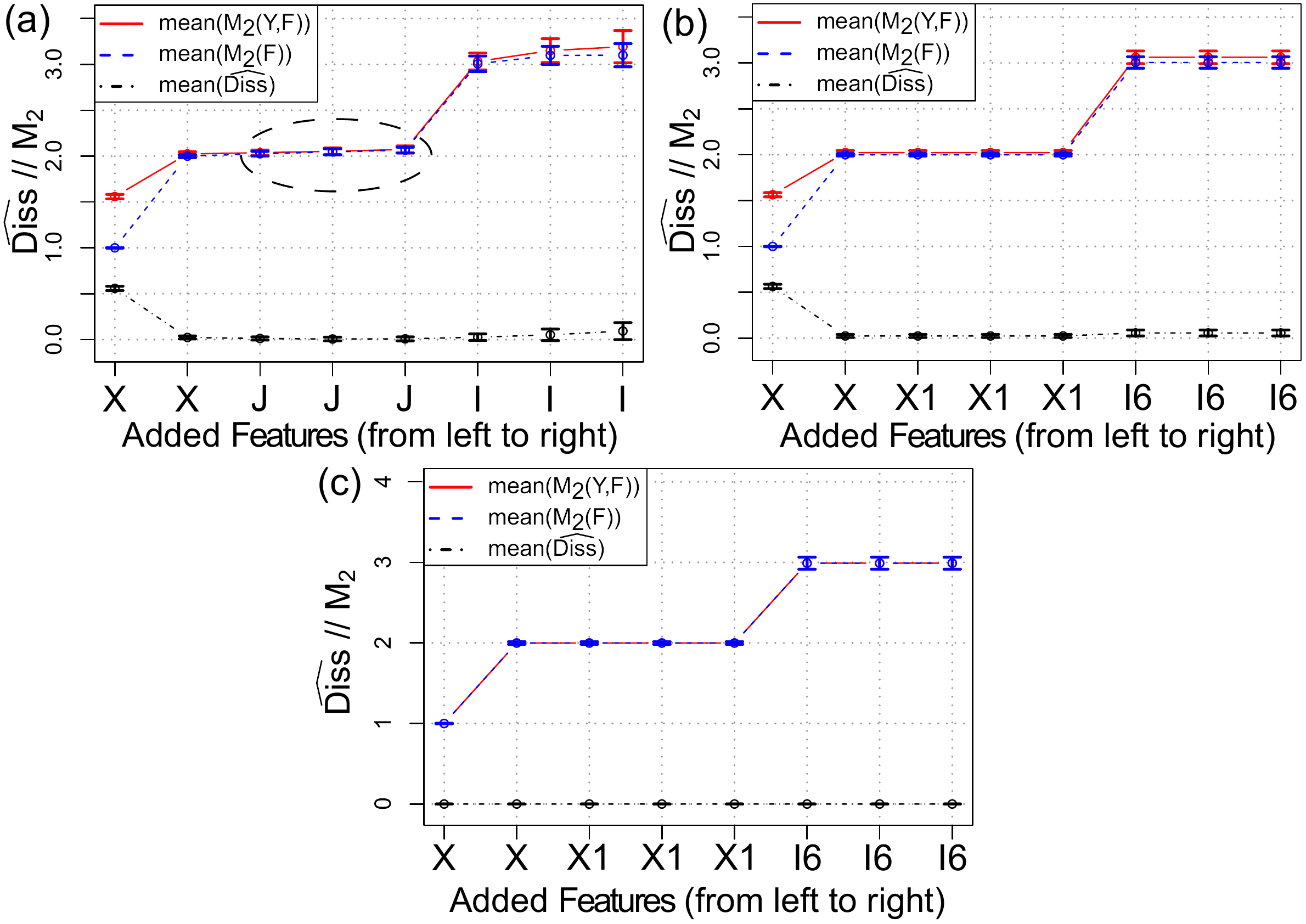}
\caption{Results of each term of Equation \ref{diss_equ_est} for (a) the original butterfly dataset ($N=1000$), for (b) a modified version in which the non-linear dependencies between the features were replaced with pure linear ones ($N=1000$), and for (c) a modified version in which $Y$ is replaced with a zero constant value ($N=1000$). Notice that the names of the features were shortened to $X$, $J$ and $I$ if the way they were ordered by the SFS search strategy changed between the simulations.}\label{Fig_Manifold}
\end{figure}

The upper panels of Figure \ref{Fig_Manifold} illustrates the impact of the non-linear dependencies on MBFR. The left-hand panel displays the results of each term of Equation \ref{diss_equ_est} for $100$ simulations of the original butterfly dataset. It can be clearly seen that the non-linearly constructed redundant and irrelevant features influence the ID estimates when they are added to the previously selected features. In contrast, the right-hand panel shows what happens when only pure linear dependencies are considered (i.e. $J_1$, $J_2$ and $J_3$ were replaced with $X_1$, $I_7$ and $I_8$ were replaced with $I_6$): the mean values and the standard deviations of the ID estimates are modified exclusively by the input variables bringing new information (either useful or useless), and the increase in the variability of the dissimilarity values is no longer progressive. It simply corresponds to $X_1$, $X_2$ and to the addition the first irrelevant feature.

There is still one aspect of the results of Figure \ref{Fig_nbr_points} which has not been fully accounted for yet: after the addition of $X_1$ and $X_2$, the dissimilarity values should be equal to $0$, but the estimates are slightly higher. Likewise, the mean values of $DR$ in Table \ref{T7} should not be lower than $1$. The observed deviations are due to the gap between $mean(M_2(F,Y))$ and $mean(M_2(F))$ that is clearly visible in the upper panels of Figure \ref{Fig_Manifold} after the addition of the first irrelevant feature. However, if the target variable $Y$ is replaced with a constant value, the gap vanishes and the dissimilarity estimates equal $0$. The last panel of Figure \ref{Fig_Manifold} shows the outcome of MBFR for such a simplified version of the butterfly manifold (again, $100$ simulations were used, and the non-linear dependencies between the input variables were replaced with pure linear ones for ease of comparison with the upper panels). This result shows that the shape and the orientation of the data manifold, along with the non-linear construction of $Y$, are key factors to explain the gap between the mean ID estimates of the original dataset. The importance of these factors might be partly related to the quadrats (i.e. the hyper-boxes) of the MINDID algorithm which cannot fit perfectly complex point patterns.

In conclusion, the complexity of the data manifolds (i.e. their shapes, their orientations, the non-linear dependencies between the features and the non-linear constructions of the output variables) affects the results by altering the terms of Equation \ref{diss_equ_est}. However, it does not prevent MBFR from identifying the relevant features.

\subsection{Redundant and Irrelevant Information}\label{info}
The MBFR algorithm aims to detect the features which are useless (i.e. irrelevant) or not necessary (i.e. redundant) to a regression problem. In addition, it is also able to help distinguish between the two types of inputs, and more precisely, between redundant and irrelevant information.

For instance, in Figure \ref{Fig_Manifold}, the first irrelevant feature causes the mean ID estimates to increase by about $1$ (i.e. by about the value of $M_2(I)$), and if it was removed, the second one would have the same effect. In contrast, the redundant features have a much smaller impact. It is even hardly noticeable for the relatively low ID values, as highlighted by the dashed ellipse in the left-hand panel. Consequently, the inputs of the butterfly data can be classified as either redundant or irrelevant according to their impacts on the ID estimates.

In real-world applications, a feature (e.g. $F_1$) rejected by MBFR could contain both redundant and irrelevant information. Nevertheless, the exact amounts of the two types of information could still be quantified by using the terms of Equation \ref{diss_equ_est}. For instance, if $F_1$ was partly redundant and partly irrelevant, it would cause an increase in the data ID which would be both higher than $0$ (fully redundant) and lower than the value of $M_2(F_1)$ (fully irrelevant). $F_1$ would also contain more irrelevant information if the increase was closer to $M_2(F_1)$ than to $0$.

In conclusion, the MBFR algorithm can help distinguish between redundant and irrelevant information by means of the ID estimates on which it relies.

\subsection{Lack of Information}\label{no_info}
This subsection investigates the behaviour of the MBFR algorithm when the relevant information is completely or partially missing.

\begin{figure}[t]
\centering
\includegraphics[width=\linewidth]{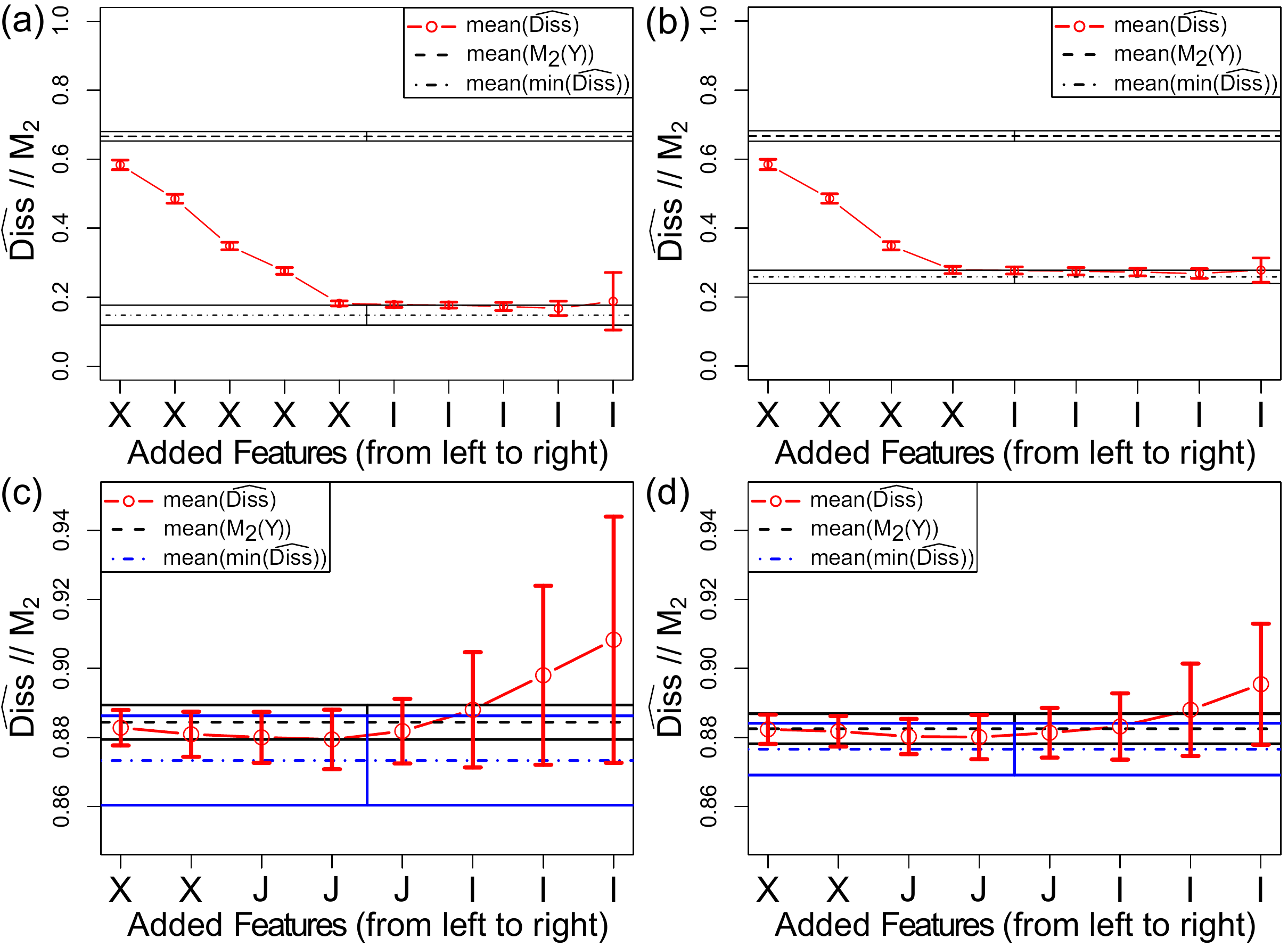}
\caption{Results of MBFR for (a) the complete Friedman dataset and (b) after the removal of $X_5$. In (c) and (d), MBFR was applied to the butterfly dataset after the shuffling of the target variable $Y$ for, respectively, $N=10000$ and $N=20000$. Notice that the names of the features were shortened to $X$, $J$ and $I$ because the way they were ordered by the SFS search strategy changed between the simulations.}\label{Fig_info}
\end{figure}

The top-left panel of Figure \ref{Fig_info} displays the results of MBFR achieved for $100$ simulations of the Friedman dataset. The algorithm distinctly detects the five relevant features and offers a clear cut-off point. The computations were rerun after the removal of $X_5$ and the results are given in the top-right panel. This time, the value of $\min(\widehat{Diss})$ is higher, and the difference accounts for the amount of information of $X_5$. This last experiment shows that the MBFR algorithm is also able to detect and quantify the absence of relevant features.

It is also worth exploring how MBFR responds to a complete absence of structure between an input and an output space. A second numerical experiment was set up to that end. The butterfly dataset was used, and $100$ simulations were generated for $N=10000$ and $N=20000$. The target variable $Y$ of each simulation was then shuffled to destroy the dependencies between the input and output spaces. The right-hand panel of Figure \ref{Fig_butterVisual} illustrates the impact of the shuffling procedure on the functional relationship between $Y$, $X_1$ and $X_2$. Finally, MBFR was applied to each simulation, and the results are displayed in the bottom panels of Figure \ref{Fig_info}. As expected, the values of $\widehat{Diss}$ stay close to $M_2(Y)$, which indicates that no features carry relevant information with regard to $Y$. The remaining gap between the mean values of $\widehat{Diss}$ and $M_2(Y)$ follows from the complexity of the data manifold discussed in Subsection \ref{manifold}, and it is more pronounced for the lower sample size, as expected from Subsection \ref{samplesize}.

Notice also that for comparison purposes, the simulations used in the lower panels of Figure \ref{Fig_info} were restricted to those for which the relevant features were selected first, followed successively by the redundant and irrelevant ones. Without this restriction, the final rankings were unpredictable, and the irrelevant features could also be picked first. This is consistent with the shuffling procedure which makes all the inputs irrelevant.

In conclusion, the MBFR algorithm provides a way to detect and quantify a lack of relevant information. In many cases, a part of that information is not truly missing, but simply corrupted with noise, and the next subsection investigates this issue.

\subsection{Additional Noise}\label{noise}
MBFR should also be able to provide reliable results when the data points are near a manifold instead of being exactly on it. This aspect was investigated by using noisy versions of the butterfly dataset. The target variable $Y$ was corrupted with a Gaussian noise (see $\varepsilon$ in Equation \ref{equ_butter}) characterized by a zero mean and a varying standard deviation (sd) ranging from $0\%$ to $100\%$ of the original standard deviation of $Y$ (the mean standard deviation of $Y$ over $100$ simulations is $0.52$). In total, seven noise thresholds were considered (i.e. $0\%$, $10\%$, $20\%$, $25\%$, $50\%$, $75\%$ and $100\%$), and for each of them the butterfly dataset was generated $100$ times with $N=10000$. The MBFR algorithm was then applied to each simulation, and the first two selected features were recorded at each run. The results are given in Table \ref{T1}.

The first two features selected by MBFR are given in the middle column, along with the number of times they were selected first over the simulations (in brackets). In addition, the minimum value of $\widehat{Diss}$, over all the simulations, is indicated in the last column. The results show that the first two features can switch position when the noise threshold is higher than $20\%$ and that $J_3$ is often substituted for $X_1$. Nevertheless, even for the greatest noise thresholds, the irrelevant features are never selected first, which means that the MBFR algorithm is still able to recognize the data manifold. Of course, the minimum dissimilarity value is higher once the noise has been added. This was to be expected according to Subsection \ref{no_info}, since the noise component partially masks the information that a variable may carry.

\begin{table}
\centering 
\footnotesize
\begin{tabular}{ccc}
\hline 
Noise Threshold    &  First Two Features (Occurrences) &  $\min(\widehat{Diss})$  \\ 
\hline 
\hline
$0\%$              & $X_1$,$X_2$ ($100$)                                       & $0.02$\\   
$10\%$             & $X_1$,$X_2$ ($100$)                                       & $0.24$\\ 
$20\%$             & $X_2$,$J_3$ ($2$) ; $X_1$,$X_2$ ($4$); $X_2$,$X_1$ ($94$) & $0.46$\\ 
$25\%$             & $X_2$,$J_3$ ($41$); $X_2$,$X_1$ ($59$)                    & $0.55$\\
$50\%$             & $X_2$,$X_1$ ($2$) ; $X_2$,$J_3$ ($98$)                    & $0.72$\\ 
$75\%$             & $X_2$,$X_1$ ($10$); $X_2$,$J_3$ ($90$)                    & $0.81$\\ 
$100\%$            & $X_2$,$X_1$ ($10$); $X_2$,$J_3$ ($90$)                    & $0.85$\\  
\hline 
\end{tabular}
\caption{The first two features selected by MBFR when applied successively to $100$ simulations of the butterfly dataset for different noise thresholds. The minimum value of $\widehat{Diss}$ over the simulations is also provided.}
\label{T1}
\end{table}

\subsection{Concluding Remarks}
The variability of the results of MBFR increases as the sample size is reduced. Besides, the use of small sample sizes highlights the impact of the manifold complexity on MBFR: MBFR is influenced by the shapes and the orientations of the data manifolds, as well as by the non-linear dependencies between the features and the non-linear constructions of the output variables. Despite this, the numerical experiments performed in this section show that MBFR is able to effectively fulfilled its goal. In other words, it can identify the relevant features and provide a clear cut-off point indicating the number of features to be retained, even in the presence of noise. MBFR also allows the user to make a distinction between redundant and irrelevant information. This is achieved by the analysis of the ID estimates involved in the implementation of the algorithm. And finally, the dissimilarity values computed from the ID estimates can help detect and quantify a possible lack of relevant information.

\section{Experimental Study Using Real Data}\label{Exp_real}
In this section, the MBFR algorithm is applied to real-world case studies from the UCI machine learning repository \cite{Lich} and the Regression website \cite{Tor}. The results are discussed with a special emphasis on the parameter $\ell^{-1}$, the coefficient of dimensional relevance $DR$, and the ability of MBFR to distinguish between redundant and irrelevant information. Finally, a comparison with a renowned filter, RReliefF \cite{Robnik03}, is conducted by using Extreme Learning Machine (ELM) \cite{Hua06}.
 
\subsection{Data}\label{subsec_real_data}
Six datasets from the UCI machine learning repository \cite{Lich} and the Regression website \cite{Tor} were used in the experiments:\begin{enumerate}[noitemsep]
\item Abalone: the goal is to predict the age of marine snails (the abalones) from physical measurements. This dataset originally contains 4177 instances and 8 features. Among the instances, two outliers were detected and removed (instances 1418 and 2052), and only the physical measurements were considered. Consequently, the resulting dataset consists of 4175 instances characterized by 7 features. 
\item Ailerons: this dataset contains 13750 instances and 40 continuous features describing the status of a F16 aircraft. The goal is to predict the control actions on the ailerons. Some issues affect the last 16 features which can be treated as either nominal or continuous. Since the decision may have an influence on the results of RReliefF, these input variables were not considered, and the dataset used in the experiments consists of 13750 instances and 24 features.
\item Boston Housing: The objective is to predict the housing prices in areas of Boston. The dataset contains 506 instances and 13 features.
\item CompAct: The goal is to predict the portion of time (\%) during which a set of CPUs run in user mode. The prediction is performed using a collection of computer system activity measures. The dataset contains 8192 instances and 21 features.
\item Parkinson's Telemonitoring: One of the objectives is to predict the motor score of the Unified Parkinson's Disease Rating Scale (UPDRS) from 16 biomedical voice measures. Telemonitoring devices were used to automatically capture speech signals in 42 patients' homes, and 5875 voice recordings were made available. One of the voice measures was removed, since it provides the same value for all the recordings, and the final dataset used in the experiments consists of 5875 instances and 15 features.
\item CT slice: the task is to predict the relative location of Computer Tomography (CT) slices on the axial axis of the human body. The prediction is carried out using features extracted from CT images. The dataset originally contains 53500 instances described by 385 features. But 63 instances are replicated several times, and 5 features provide a constant value. Consequently, the data preprocessing resulted in a slightly modified dataset consisting of 53436 instances and 380 features.
\end{enumerate}

\subsection{Feature Selection with MBFR}
The MBFR algorithm was applied to the real world datasets described in the previous subsection. For each of them, the set of values of the parameter $\ell^{-1}$ was chosen as follows:\begin{enumerate}[noitemsep]
\item The plot relating $\ln(I_{m=2,\ell^{-1}})$ to $\ln(\ell^{-1})$ was computed for the full dataset (including all the instances, all the features and the target variable) with $\ell^{-1} \in \lbrace 1,2,\ldots,130\rbrace$.
\item The upper and lower bounds of the set were given by the extent of the linear part of the plot. For most of the datasets, the upper bound turned out to be simply the maximum value of $\ell^{-1}$ ensuring the presence of two points in, at least, one quadrat.
\item If the upper bound was lower than 30, every integer value within the bounds was retained. But, if it was equal to or higher than 30, only the integer values following a geometric progression with ratio $2$ were used, and the bounds were modified accordingly. This allows the MBFR algorithm to run faster by reducing the value of $R$ (see Section \ref{Mindex}).
\end{enumerate}

The resulting sets of values of the parameter $\ell^{-1}$ are given in Table \ref{T3}. Although it might seem better to change the values for each feature combination, it turned out not to be necessary. Once a set had been built by following the described procedure, it was used throughout the feature selection process. The CompAct dataset was the only exception: the log-log plots of several single features were characterized by two distinct linear parts. The steepest one was retained, since it led to a higher value of $DR$. Finally, Table \ref{T3} also gives the values $M_2$ for the whole datasets (including the target variables). These values suggest that the dimensions of the spaces in which the data points truly reside could be smaller than that of the original data spaces. In other words, the datasets could contain redundant information that MBFR might uncover.

\begin{table}
\centering 
\footnotesize
\begin{tabular}{ccccccc}
\hline
Datasets &  Parameter $\ell^{-1}\in$            &  $M_2$  &\# F.  &  $\widehat{Diss}$ &   $DR$\\ 
\hline 
\hline
Abalone  & $\lbrace 4,8,16,32,64\rbrace$        & $3.66$  &$3 (7)$     &     $0.36$       & $0.46$ \\   
Ailerons & $\lbrace 5,6,\ldots,24,25\rbrace$    & $5.34$  &$7 (24)$    &     $0.22$       & $0.71$ \\ 
Housing  & $\lbrace 2,3,\ldots,18,19\rbrace$    & $3.30$  &$8 (13)$    &     $0.14$       & $0.84$ \\
CompAct  & $\lbrace 1,2,\ldots,9,10\rbrace$     & $2.16$  &$6 (21)$    &     $0.03$       & $0.94$ \\
Parkinson& $\lbrace 1,2,4,8,16,32,64\rbrace$    & $4.30$  &$8 (15)$    &     $0.58$       & $0.31$ \\
CT slice & $\lbrace 1,2,4,8,16,32,64,128\rbrace$& $3.09$  &$22 (380)$  &  $\approx 0.00$  & $\approx 1.00$ \\
\hline 
\end{tabular}
\caption{Parameters and Results of the application of MBFR to the real world datasets. The values in brackets in the $4^{th}$ column are the total numbers of features in the datasets, and ``\# F.'' stands for ``number of selected features''. Besides, $\widehat{Diss}$ and $DR$ were computed by considering only the selected features, while $M_2$ is given for the whole datasets (including the target variables).}
\label{T3}
\end{table}

\begin{figure}
\centering
\includegraphics[width=\linewidth]{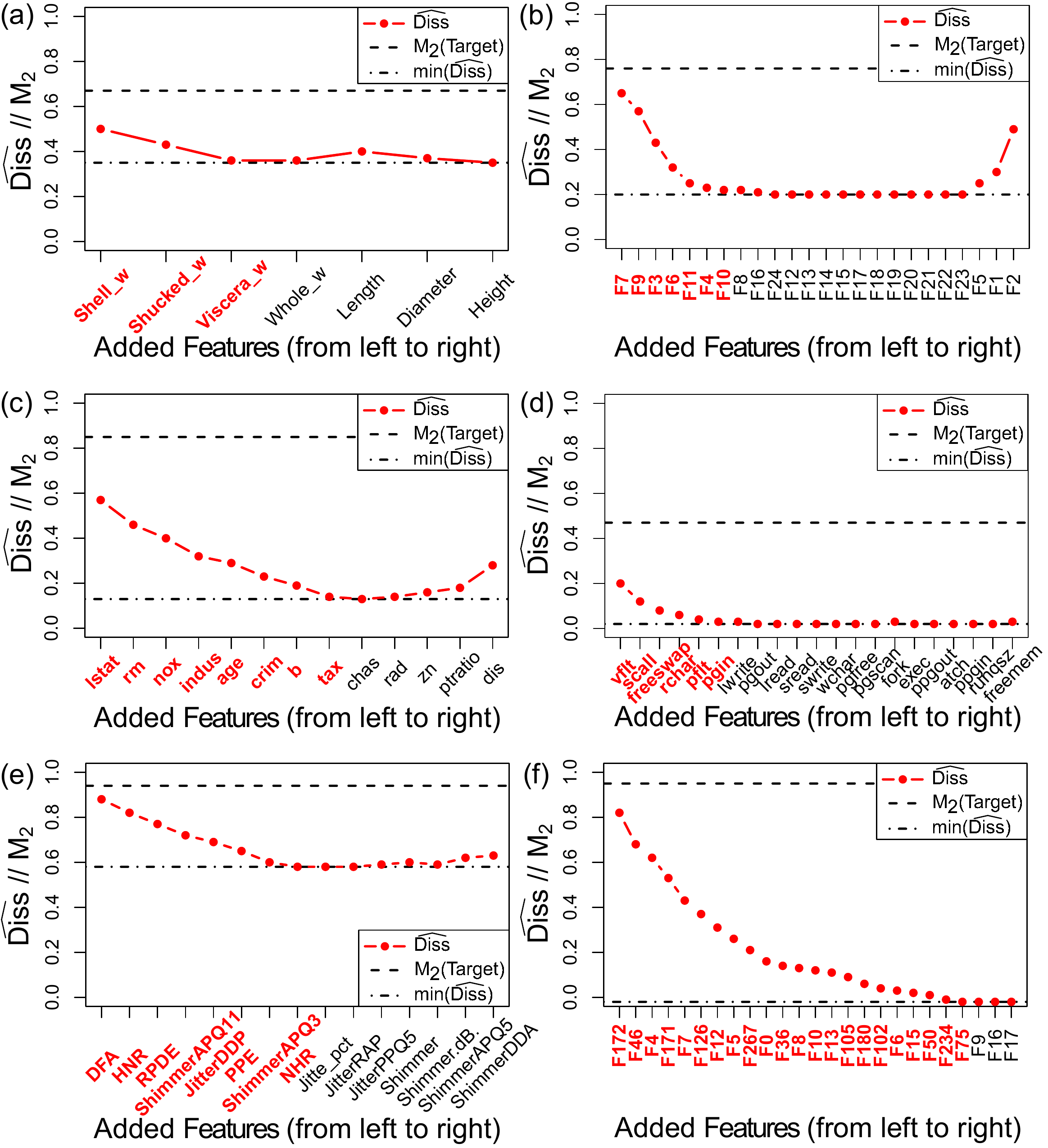}
\caption{Results of MBFR: (a) Abalone (b) Ailerons (c) Boston Housing (d) CompAct, (e) Parkinson's Telemonitoring, (f) CT slice. The selected features are indicated in bold red font, and ``Target'' refers to the output (or target) variable of a dataset.}\label{RealData_MBFR}
\end{figure}

The MBFR algorithm was applied to each dataset with the parameter values of Table \ref{T3}. The results are displayed in Figure \ref{RealData_MBFR}. In each panel, a relatively clear cut-off point allows the user to identify the features to be selected. Moreover, Table \ref{T3} summarizes the results by providing the dissimilarity estimates (i.e. $\widehat{Diss}$) and the values of $DR$ that were computed by considering only the selected features. This overview shows that MBFR leads to a significant reduction in the number of features and that the chosen datasets cover a wide range of situations. For instance, in the Parkinson dataset, 8 features out of 15 are detected as relevant, and they account for about a third of the information contained in the target variable. In contrast, in the CT slice dataset, 359 features are considered redundant or irrelevant, and the relevant ones fully explain the relative location of the CT slices (i.e the target variable).

In conclusion, this subsection highlights the effectiveness of the MBFR algorithm for feature selection in real world applications. In the next subsection, the results will be validated by means of a comparison with a benchmark algorithm called RReliefF.

\subsection{Comparison with RReliefF Using Extreme Learning Machine}
In this subsection, the performance of MBFR is compared with that of RReliefF \cite{Robnik96,Robnik97,Robnik03} using a comprehensive evaluation procedure based on Extreme learning Machine (ELM) \cite{Hua06,GHua15}. A brief introduction to ELM and the Relief family of algorithms is also provided for a good understanding of the results. 

The Relief family comprises three main algorithms for feature selection: Relief \cite{Kir92}, ReliefF \cite{Kon94,Kon97} and RReliefF \cite{Robnik96,Robnik97,Robnik03}. They all consist in attributing scores to the features according to how well their values can distinguish between instances that are close to one another in the data space. Relief achieves this goal for two-class classification problems. It randomly selects an instance and searches for its two nearest neighbours: one from the same class and one from the other class. After that, it updates the scores of the features depending on their values for the randomly selected instance and the two neighbours. The operation is repeated $m_{RF}$ times, and a final score lower than $0$ indicates that a feature might be irrelevant. This threshold, determining whether a feature should be retained, is one of the major advantages of the technique. Following a similar procedure, the ReliefF algorithm is an extension which can deal with multiclass problems and with incomplete and noisy data.

RReliefF (i.e. Regressional ReliefF) is an adaptation of Relief and ReliefF to regression. In regression, the exact knowledge of whether two instances belong to the same class or not cannot be used. RReliefF replaces it with a probability that the predicted values of two instances will be different. Moreover, it computes the final score of each feature by taking into account $k_{RF}$ neighbours. Among these neighbours, the closest ones should have a greater influence, and a kernel of parameter $\sigma_{RF}$ can be used to assign a weight to each of them.  

The algorithms of the Relief family have often been used as benchmarks \cite{Hall00,Sousa07,Vi16}. In this research, RReliefF was applied to the real world datasets of Subsection \ref{subsec_real_data}, with the exception of the CT slice data that contain too many instances. The R package ``CORElearn'' \cite{Core15} was used with the evaluation heuristic ``RReliefFexpRank'' and its default parameters: $m_{RF}=N$ ($N$ is the number of instances in the datasets), $k_{RF}=70$ and $\sigma_{RF}=20$. These parameters were tested and turned out to be suitable for each dataset. Besides, RReliefF was run with two relevance thresholds: $0.00$ and $0.01$ as suggested in \cite{Hall00}. Any feature with a final score less than the specified threshold was considered irrelevant. 
  
Table \ref{T5} compares the ability of RReliefF and MBFR to reduce the dimensionality of the datasets. In all but one case, MBFR performs more feature selection than RReliefF with the relevance threshold of $0.00$. The differences are less pronounced with the relevance threshold of $0.01$ for which the two algorithms achieve comparable results on three datasets. Nevertheless, they still provide distinct outcomes for the CompAct and Parkinson data, and they never select the exact same features. Consequently, the comparison requires a way of assessing the amount of information contained in the selected features. This is the reason why ELM is used in this paper.

ELM is a single layer feed-forward neural network which can achieve the same accuracy as the well-known Multilayer Perceptrons (MLP) \cite{Ros58,Ros62,Wer74,Rum86}, while being much faster. Its main specificity lies in the weights connecting the inputs to the hidden layer. They are randomly generated and never updated, and then the weights between the hidden layer and the outputs are learnt in a single step. In this way, the number $\tilde{N}$ of hidden nodes is the only hyperparameter of ELM, making its implementation rather straightforward. However, an activation function is also required and a sigmoid function was used in this paper.

ELM has been applied successfully in a wide range of case studies \cite{Bar13,Leu14,Me15}, and its high efficiency makes it well-suited to carry out comparisons between feature selection techniques. It is also worth mentioning that RReliefF and ELM have already been combined to effectively improve learning performance in temperature forecasting \cite{Mat13}. In this paper, ELM was used to evaluate the predictive power of the subsets of features selected by MBFR and RReliefF. This evaluation was achieved according to a procedure which was partly presented in \cite{Ja97,Reu03} to prevent overfitting \cite{Li16}. It can be subdivided into 5 steps:\begin{enumerate}[noitemsep]
\item $20\%$ of the $N$ instances are randomly assigned to a test set, and the remaining $80\%$ are passed on to Step 2. The same split is used for all the subsets of features being compared.
\item The data are projected into the $[0,1]$ interval, and the coefficients of the projection are recorded.
\item \label{step_cv} The value of the hyperparameter $\tilde{N}$ is selected by performing 10-fold cross-validation. For each value of $\tilde{N}\in\lbrace1,\ldots,350\rbrace$, 1 fold is iteratively allocated to a set of validation, and the remaining 9 folds are assigned to a training set. For each of the 10 iterations, an ELM model is fit on the training observations, and the Mean Square Error (MSE) is computed using the validation set. Then the 10 MSEs are averaged to provide an estimate of the true error, and their standard deviation is recorded. Finally, the value of $\tilde{N}$ resulting in the lowest error estimate is retained for Step 4, unless the corresponding variability is too high. In that case, $\tilde{N}$ is manually tuned to find a good compromise between the mean and the variance of the error.
\item Using all the instances involved in the cross-validation procedure (i.e. $80\%$ of the original data), a new model is trained with the value of $\tilde{N}$ from Step 3. Then a prediction is made for the instances of the test set (i.e. $20\%$ of the original data) after they have been projected to the $[0,1]$ interval using the coefficient of Step 2. The operation (training and prediction) is repeated 100 times to account for the variability of the weights connecting the inputs to the hidden nodes. The final prediction for each instance is computed by averaging the 100 values and by rescaling the results to the original output range (using coefficients of Step 2). Finally, the relative mean squared error \cite{Robnik96} is calculated on the test set. It is defined as follows:\begin{equation}\label{Eq_RE} 
RE_{tst}=\frac{\sum_{i=1}^{N_{tst}}(y_i-\hat{y}_i)^2}{\sum_{i=1}^{N_{tst}}(y_i-\bar{y}_{tst})^2}
\end{equation}where $N_{tst}$ is the number of instances in the test set, $y_i$ is the measurement of the output variable for the $i^{th}$ instance, $\hat{y}_i$ is the corresponding prediction, and $\bar{y}_{tst}$ is the mean of the output variable computed on the test set. Lower values of $RE_{tst}$ are better and a value higher than $1$ indicates that the tested model performs worse than the mean. 
\item Steps 1 to 4 are repeated 20 times to account for the randomness in the data splits of step 1. After that, the mean and standard deviation of the 20 values of $RE_{tst}$ are calculated and are used to assess the predictive power of the selected features. 
\end{enumerate}

For each dataset, four sets of features were passed on to the evaluation procedure: the set selected by MBFR, the two sets selected by RReliefF (one for each relevance threshold) and a benchmark set selected by a technique called ELM\_{SFS}. ELM\_{SFS} is a simple wrapper approach combining ELM and the same SFS search strategy as MBFR. It works as follows: at each step of the search process, the predictive power of each set of features is assessed using the same cross-validation as in Step \ref{step_cv} of the evaluation procedure; and finally, the set returning the lowest MSE over the entire SFS is selected. 

Table \ref{T5} presents the results of the evaluation procedure for each feature selection technique. The mean values of $RE_{tst}$ over the 20 iterations are provided, along with the corresponding standard deviations. The sets of features selected by MBFR provide comparable or better accuracies (i.e. lower values of $RE_{tst}$) than those resulting from RReliefF. This is true even when MBFR performs more feature selection. Besides, although neither RReliefF nor MBFR improves the performance of ELM, only MBFR is able to maintain or stay close to (i.e. no more than one standard deviation away from) the mean values of $RE_{tst}$ achieved by ELM\_{SFS} for each dataset. 

\begin{table}
\centering 
\scriptsize
\begin{tabular}{ccccccccc}
\cline{2-9}
&\multicolumn{2}{c}{RReliefF 0.00}&\multicolumn{2}{c}{RReliefF 0.01}&\multicolumn{2}{c}{MBFR}&\multicolumn{2}{c}{ELM\_{SFS}}\\
\hline 
Datasets &\# F.&   $RE_{tst}$   & \# F.&    $RE_{tst}$    & \# F. &   $RE_{tst}$ & \# F.    &  $RE_{tst}$     \\
\hline 
\hline
Abalone  & $7^{\ast}$& {\boldmath$0.43$($0.02$)}  &$1$ &$0.57$($0.02$)            &$3$ & \underline{$0.46$($0.03$)} &$5$ & {\boldmath$0.43$($0.02$)}\\   
Ailerons &$24^{\ast}$& {\boldmath$0.15$($0.01$)}  &$8$ &$0.24$($0.01$)            &$7$ & {\boldmath$0.15$($0.01$)} &$9$& {\boldmath$0.15$($0.01$)}\\
Housing  &$10$&  $0.17$($0.06$) &$8$ &$0.20$($0.06$) &$8$ & \underline{$0.16$($0.06$)} &$9$& {\boldmath$0.13$($0.05$)}\\ 
CompAct  &$21^{\ast}$& {\boldmath$0.02$($0.00$)}  &$20$ &{\boldmath$0.02$($0.00$)} &$6$ & {\boldmath$0.02$($0.00$)} &$11$& {\boldmath $0.02$($0.00$)}\\ 
Parkinson& $2$&  $0.90$($0.02$) &$1$ &$0.98$($0.01$) &$8$ & \underline{$0.81$($0.02$)} &$6$& {\boldmath$0.79$($0.03$)}\\ 
\hline 
\end{tabular}
\caption{Comparison between RReliefF and MBFR based on the number of selected features (i.e. \# F.) and the relative mean squared error $RE_{tst}$. The table provides the mean values and the standard deviations (in brackets) of $RE_{tst}$ over the 20 iterations of the evaluation procedure. And the number of selected features is marked with an asterisk if the feature selection process did not lead to a reduction in the dimensionality of the data. Moreover, the best results are indicated in bold script (the lower, the better) and the second best results are underlined.}
\label{T5}
\end{table}

In terms of computing time, MBFR turned out to be competitive with RReliefF, since the two algorithms run in less than 30 seconds (s) on all datasets, with the exception of Ailerons. On the Ailerons dataset, MBFR was slower than RReliefF. It performed feature selection in about 220 s, while only 56 s were necessary for RReliefF. This significant difference is mainly due to the fact that Ailerons is the dataset requiring the largest number of values of the parameter $\ell^{-1}$. Despite this, MBFR is extremely fast for an algorithm following a SFS search strategy. By comparison, ELM\_{SFS} took more than 100 hours to complete the full feature selection procedure, while leading to the same mean error values as MBFR. Notice that all the numerical experiments were carried out using an Intel Core i7-2600 CPU @ 3.40 GHz along with 16.0 GB of RAM under Windows 7, and the value of $C$ was set to $E-1$ (i.e. the default value) for each dataset.

Another interesting point is the relationship between $RE_{tst}$ and the coefficient of dimensional relevance $DR$. The correlation between the two measures was computed with the mean values of $RE_{tst}$ resulting from MBFR and the values of $DR$ given in Table \ref{T3}. It turned out that Pearson's coefficient was equal to $-0.96$, which tends to confirm that $DR$ is a promising measure of feature relevance.

\section{Conclusion}\label{End}
This paper presents a new algorithm for supervised feature selection, namely the Morisita-Based Filter for Regression problems (MBFR). As its name suggests, it is designed for regression problems, and it relies on the recently introduced Morisita estimator of Intrinsic Dimension (ID). Comprehensive numerical experiments were carried out using two simulated datasets: the well-known Friedman dataset and the butterfly dataset which was specifically designed for the needs of this research. Different sample sizes, noise levels and non-linear dependencies were tested, and the variability of the results was examined by means of Monte Carlo simulations. MBFR was shown to be an effective tool for reducing the dimensionality of large datasets of varying complexity. Besides, the ability of the algorithm to distinguish between redundant and irrelevant information was presented and successfully tested. 

MBFR was applied to real world datasets from publicly accessible repositories. An innovative methodology was implemented to conduct a comparison with a benchmark algorithm called RReliefF. MBFR resulted in better or comparable performance according to the accuracy achieved by Extreme Learning Machine (ELM). This was true even when MBFR retained fewer features than RReliefF.

A new coefficient of relevance was introduced, namely the coefficient of dimensional relevance $DR$. It was estimated by using MBFR, and its reliability was evaluated by means of ELM. $DR$ is exclusively based on the ID concept, it is easily interpretable, and it can be applied to high-dimensional datasets.

Finally, this paper shows that ID-based methods have the potential to improve the performance of existing machine learning algorithms. In addition to the presented work, they can also contribute to the development of new powerful tools to conduct fundamental tasks, such as classification, clustering and pattern detection.

\section{Acknowledgements}
The authors are grateful to the anonymous reviewers for their helpful and constructive comments that contributed to improving the paper. They also would like to thank Mohamed Laib and Zhivko Taushanov for many fruitful discussions about machine learning and statistics.

\bibliographystyle{elsarticle-num}
\bibliography{References}

\begin{thebibliography}{10}
\expandafter\ifx\csname url\endcsname\relax
  \def\url#1{\texttt{#1}}\fi
\expandafter\ifx\csname urlprefix\endcsname\relax\def\urlprefix{URL }\fi
\expandafter\ifx\csname href\endcsname\relax
  \def\href#1#2{#2} \def\path#1{#1}\fi

\bibitem{Bell61}
R.~Bellman, Adaptive Control Processes: A Guided Tour, Princeton University
  Press, Princeton (US-NJ), 1961.

\bibitem{Guy03}
I.~Guyon, A.~Elisseeff, An introduction to variable and feature selection,
  Journal of Machine Learning Research 3 (2003) 1157--1182.

\bibitem{Guy06}
I.~Guyon, S.~Gunn, M.~Nikravesh, L.~A. Zadeh (Eds.), Feature {E}xtraction:
  {F}oundations and {A}pplications, Springer, Berlin, 2006.

\bibitem{Ghe10}
I.~A. Gheyas, L.~S. Smith, Feature subset selection in large dimensionality
  domains, Pattern Recognition 43~(1) (2010) 5--13.

\bibitem{Zen15}
Z.~Zeng, H.~Zhang, R.~Zhang, C.~Yin, A novel feature selection method
  considering feature interaction, Pattern Recognition 48 (2015) 2656--2666.

\bibitem{Robnik03}
M.~Robnik-\v{S}ikonja, I.~Kononenko, Theoretical and {E}mpirical {A}nalysis of
  {R}elief{F} and {RR}elief{F}, Machine Learning 53~(1) (2003) 23--69.

\bibitem{Pen05}
H.~C. Peng, F.~Long, C.~Ding, Feature selection based on mutual information
  criteria of max-dependency, max-relevance, and min-redundancy, IEEE
  Transactions on Pattern Analysis and Machine Intelligence 27~(8) (2005)
  1226--1238.

\bibitem{Hall00}
M.~A. Hall, Correlation-based feature selection for discrete and numeric class
  machine learning, in: Proceedings of the 17th International Conference on
  Machine Learning (ICML), Stanford (USA), 2000.

\bibitem{Koh97}
R.~Kohavi, G.~H. John, Wrappers for feature subset selection, Artificial
  Intelligence 97~(1-2) (1997) 273--324.

\bibitem{Leu14}
M.~Leuenberger, M.~Kanevski, Feature selection in environmental data mining
  combining simulated annealing and extreme learning machine, in: Proceedings
  of the 22nd European Symposium on Artificial Neural Networks, Computational
  Intelligence and Machine Learning (ESANN), d-side pub., 2014, pp. 601--606.

\bibitem{Tib96}
R.~Tibshirani, Regression shrinkage and selection via the {L}asso, Journal of
  the Royal Statistical Society, Series B 58~(1) (1996) 267--288.

\bibitem{Bre01}
L.~Breiman, Random forests, Machine Learning 45~(1) (2001) 5--32.

\bibitem{Cot01}
S.~F. Cotter, K.~Kreutz-Delgado, B.~D. Rao, Backward sequential elimination for
  sparse vector subset selection, Signal Processing 81~(9) (2001) 1849--1864.

\bibitem{Col03}
S.~Colak, C.~Isik, Feature subset selection for blood pressure classification
  using orthogonal forward selection, in: Proceedings of the 29th IEEE Annual
  Bioengineering Conference, 2003, pp. 122--123.

\bibitem{Whit71}
A.~W. Whitney, A direct method of nonparametric measurement selection, IEEE
  Transactions on Computers 20~(9) (1971) 1100--1103.

\bibitem{Mei06}
R.~Meiri, J.~Zahavi, Using simulated annealing to optimize the feature
  selection problem in marketing applications, European Journal of Operational
  Research 171~(3) (2006) 842--858.

\bibitem{Kirk83}
S.~Kirkpatrick, C.~D. Gelatt, M.~P. Vecchi, Optimization by simulated
  annealing, Science 220~(4598) (1983) 671--680.

\bibitem{Ta15}
S.~Tabakhi, P.~Moradi, Relevance–redundancy feature selection based on ant
  colony optimization, Pattern Recognition 48~(9) (2015) 2798--2811.

\bibitem{Do92}
M.~Dorigo, Optimization, learning and natural algorithms, Ph.D. thesis,
  Politecnico di Milano (1992).

\bibitem{Rob13}
S.~Robert, L.~Foresti, M.~Kanveski, Spatial prediction of monthly wind speeds
  in complex terrain with adaptive general regression neural networks,
  International Journal of Climatology 33~(7) (2013) 1793--1804.

\bibitem{Go15Esann}
J.~Golay, M.~Leuenberger, M.~Kanevski, Morisita-based feature selection for
  regression problems, in: Proceedings of the 23rd European Symposium on
  Artificial Neural Networks, Computational Intelligence and Machine Learning
  (ESANN), d-side pub., 2015, pp. 279--284.

\bibitem{Cama03}
F.~Camastra, Data dimensionality estimation methods: a survey, Pattern
  Recognition 36~(12) (2003) 2945--2954.

\bibitem{LeeVer07}
J.~A. Lee, M.~Verleysen, Nonlinear Dimensionality Reduction, Springer,
  New-York, 2007.

\bibitem{Cama16}
F.~Camastra, A.~Staiano, Intrinsic dimension estimation: Advances and open
  problems, Information Sciences 328 (2016) 26--41.

\bibitem{Go15}
J.~Golay, M.~Kanevski, A new estimator of intrinsic dimension based on the
  multipoint {M}orisita index, Pattern Recognition 48~(12) (2015) 4070--4081.

\bibitem{Trai00}
C.~Traina~Jr., A.~J.~M. Traina, L.~Wu, C.~Faloutsos, Fast feature selection
  using fractal dimension, in: Proceedings of the XV Brazilian Symposium on
  Databases (SBBD), 2000, pp. 158--171.

\bibitem{Hua06}
G.-B. Huang, Q.-Y. Zhu, C.-K. Siew, Extreme learning machine: Theory and
  applications, Neurocomputing 70~(1-3) (2006) 489--501.

\bibitem{Ek11}
B.~Eriksson, M.~Crovella, Estimation of intrinsic dimension via clustering,
  Tech. rep., Boston University, Department of Computer Science (2011).

\bibitem{Go14}
J.~Golay, M.~Kanevski, C.~D. Vega~Orozco, M.~Leuenberger, The multipoint
  {M}orisita index for the analysis of spatial patterns, Physica A 406 (2014)
  191--202.

\bibitem{Grass833}
P.~Grassberger, I.~Procaccia, Measuring the strangeness of strange attractors,
  Physica D 9~(1-2) (1983) 189--208.

\bibitem{Bor93}
S.~Borgani, G.~Murante, A.~Provenzale, R.~Valdarnini, Multifractal analysis of
  the galaxy distribution: {R}eliability of results from finite data sets,
  Physical Review E 47~(6) (1993) 3879--3888.

\bibitem{lov87}
S.~Lovejoy, D.~Schertzer, A.~Tsonis, Functional box-counting and multiple
  elliptical dimensions in rain, Science 235~(4792) (1987) 1036--1038.

\bibitem{Hua94}
Q.~Huang, J.~R. Lorch, R.~C. Dubes, Can the fractal dimension of images be
  measured?, Pattern Recognition 27~(3) (1994) 339--349.

\bibitem{Xu15}
Y.~Xu, Y.~Quan, Z.~Zhang, H.~Ling, H.~Ji, Classifying dynamic textures via
  spatiotemporal fractal analysis, Pattern Recognition 48~(10) (2015)
  3239--3248.

\bibitem{Mand83}
B.~B. Mandelbrot, {T}he {F}ractal {G}eometry of {N}ature, W.H. Freeman, San
  Francisco, 1983.

\bibitem{Ott93}
E.~Ott, Chaos in dynamical systems, Cambridge University Press, Cambridge (UK),
  1993.

\bibitem{Hent83}
H.~G.~E. Hentschel, I.~Procaccia, The infinite number of generalized dimensions
  of fractals and strange attractors, Physica D 8~(3) (1983) 435--444.

\bibitem{Mo12}
D.~Mo, S.~H. Huang, Fractal-based intrinsic dimension estimation and its
  application in dimensionality reduction, IEEE Transactions on Knowledge and
  Data Engineering 24~(1) (2012) 59--71.

\bibitem{Trai10}
C.~Traina~Jr., A.~J.~M. Traina, C.~Faloutsos, Fast feature selection using
  fractal dimension - {T}en years later, Journal of Information and Data
  Management 1~(1) (2010) 17--20.

\bibitem{Sousa07}
E.~P.~M. De~Sousa, C.~Traina~Jr., A.~J.~M. Traina, L.~Wu, C.~Faloutsos, A fast
  and effective method to find correlations among attributes in databases, Data
  Mining and Knowledge Discovery 14~(3) (2007) 367--407.

\bibitem{Lee06}
H.~D. Lee, M.~C. Monard, F.~C. Wu, A fractal dimension based filter algorithm
  to select features for supervised learning, in: J.~S. Sichman, H.~Coelho,
  S.~O. Rezende (Eds.), Advances in Artificial Intelligence - IBERAMIA-SBIA
  2006, Springer, 2006, pp. 278--288.

\bibitem{Pham09}
D.~T. Pham, M.~S. Packianather, M.~S. Garcia, M.~Castellani, Novel feature
  selection method using mutual information and fractal dimension, in:
  Proceedings of the 35th Annual Conference of IEEE on Industrial Electronics
  (IECON), 2009, pp. 3393--3398.

\bibitem{Hul90}
S.~H. Hurlbert, Spatial {D}istribution of the {M}ontane {U}nicorn, Oikos 58~(3)
  (1990) 257--271.

\bibitem{Mori59}
M.~Morisita, {M}easuring of the {D}ispersion of {I}ndividuals and {A}nalysis of
  the {D}istributional {P}atterns, {M}emoires of the {F}aculty of {S}cience
  ({S}erie {E}), {K}yushu {U}niversity 2~(4) (1959) 215--235.

\bibitem{Rcran}
{R Development Core Team}, \href{http://www.R-project.org}{R: A Language and
  Environment for Statistical Computing}, R Foundation for Statistical
  Computing, Vienna, Austria, {ISBN} 3-900051-07-0 (2008).
\newline\urlprefix\url{http://www.R-project.org}

\bibitem{Fri91}
J.~H. Friedman, Multivariate {A}daptive {R}egression {S}plines, The Annals of
  Statistics 19~(1) (1991) 1--67.

\bibitem{Tor}
L.~Torgo,
  \href{http://www.dcc.fc.up.pt/$\sim$ltorgo/Regression/DataSets.html}{{R}egression
  {D}ata{S}ets}.
\newline\urlprefix\url{http://www.dcc.fc.up.pt/$\sim$ltorgo/Regression/DataSets.html}

\bibitem{Lich}
M.~Lichman, \href{http://archive.ics.uci.edu/ml}{{UCI} {M}achine {L}earning
  {R}epository}, University of California, Irvine, School of Information and
  Computer Sciences (2013).
\newline\urlprefix\url{http://archive.ics.uci.edu/ml}

\bibitem{Robnik96}
M.~Robnik-\v{S}ikonja, I.~Kononenko, Context sensitive attribute estimation in
  regression, in: Proceedings of the ICML-96 Workshop on Learning in
  Context-Sensitive Domains, Bari (IT), 1996, pp. 43--52.

\bibitem{Robnik97}
M.~Robnik-\v{S}ikonja, I.~Kononenko, An adaptation of {R}elief for attribute
  estimation in regression, in: Proceedings of the 14th International
  Conference on Machine Learning (ICML), Nashville (USA), 1997, pp. 296--304.

\bibitem{GHua15}
G.~Huang, G.-B. Huang, S.~Song, K.~You, Trends in extreme learning machines: A
  review, Neural Networks 61 (2015) 32--48.

\bibitem{Kir92}
K.~Kira, L.~A. Rendell, The feature selection problem: Traditional methods and
  a new algorithm, in: Proceedings of the 10th national conference on
  Artificial intelligence, San Jose (US-CA), 1992, pp. 129--134.

\bibitem{Kon94}
I.~Kononenko, Estimating attributes: analysis and extensions of relief, in:
  F.~Bergadano, L.~De~Raedt (Eds.), Machine Learning: ECML-94, Springer, 2006,
  pp. 171--182.

\bibitem{Kon97}
I.~Kononenko, E.~\v{S}imec, M.~Robnik-\v{S}ikonja, Overcoming the myopia of
  inductive learning algorithms with {RELIEFF}, Applied Intelligence 7~(1)
  (1997) 39--45.

\bibitem{Vi16}
N.~X. Vinh, S.~Zhou, J.~Chan, J.~Bailey, Can high-order dependencies improve
  mutual information based feature selection?, Pattern Recognition 53 (2016)
  46--58.

\bibitem{Core15}
M.~Robnik-\v{S}ikonja, P.~Savicky, J.~Adeyanju~Alao,
  \href{http://CRAN.R-project.org/package=CORElearn}{CORElearn: Classification,
  Regression and Feature Evaluation}, {R} package version 0.9.45 (2015).
\newline\urlprefix\url{http://CRAN.R-project.org/package=CORElearn}

\bibitem{Ros58}
F.~Rosenblatt, The perceptron: a probabilistic model for information storage
  and organization in the brain, Psychological Review 65~(6) (1958) 386--408.

\bibitem{Ros62}
F.~Rosenblatt, Principles of Neurodynamics: Perceptrons and the Theory of Brain
  Mechanisms, Spartan, Washington (DC), 1962.

\bibitem{Wer74}
P.~Werbos, Beyond regression: New tools for prediction and analysis in the
  behavioral sciences, Ph.D. thesis, Harvard University (1974).

\bibitem{Rum86}
D.~Rumelhart, G.~Hinton, R.~Williams, Learning internal representations by
  error propagation, in: D.~Rumelhart, J.~McClelland (Eds.), Parallel
  Distributed Processing: Explorations in the Microstructure of Cognition,
  Vol.~1, The MIT Press, Cambridge (USA), 1986, pp. 318--362.

\bibitem{Bar13}
A.~Baradarani, Q.~M.~J. Wu, M.~Ahmadi, An efficient illumination invariant face
  recognition framework via illumination enhancement and {DD-DTCWT} filtering,
  Pattern Recognition 46~(1) (2013) 57--72.

\bibitem{Me15}
J.~J. De~Mesquita S\'{a}~Junior, A.~R. Backes, {ELM} based signature for
  texture classification, Pattern Recognition 51 (2016) 395--401.

\bibitem{Mat13}
F.~Mateo, J.~J. Carrasco, M.~Mill{\'a}n-Giraldo, A.~Sellami,
  P.~Escandell-Montero, J.~M. Mart{\'i}nez-Mart{\'i}nez, E.~Soria-Olivas,
  Temperature forecast in buildings using machine learning techniques, in:
  Proceedings of the 21st European Symposium on Artificial Neural Networks,
  Computational Intelligence and Machine Learning (ESANN), d-side pub., 2013,
  pp. 357--362.

\bibitem{Ja97}
A.~Jain, D.~Zongker, Feature selection: Evaluation, application, and small
  sample performance, IEEE Transactions on Pattern Analysis and Machine
  Intelligence 19~(2) (1997) 153--158.

\bibitem{Reu03}
J.~Reunanen, Overfitting in making comparisons between variable selection
  methods, Journal of Machine Learning Research 3 (2003) 1371--1382.

\bibitem{Li16}
R.~Liu, D.~F. Gillies, Overfitting in linear feature extraction for
  classification of high-dimensional image data, Pattern Recognition 53 (2016)
  73--86.

\end{thebibliography}
 
\end{document}